\pdfoutput=1
\documentclass[11pt]{article}
\usepackage[table]{xcolor}

\usepackage{acl}
\usepackage{listings}
\usepackage{booktabs}

\usepackage{times}
\usepackage{latexsym}
\usepackage{multirow}

\usepackage[T1]{fontenc}

\usepackage[utf8]{inputenc}

\usepackage{microtype}

\usepackage{inconsolata}

\usepackage{graphicx}

\usepackage{xspace}
\usepackage{booktabs}

\usepackage{tcolorbox}
\usepackage{colortbl}

\definecolor{boxgray}{gray}{0.95}

\newcommand{\IT}{$^{\clubsuit}$}
\newcommand{\ist}{$^{\spadesuit}$}
\newcommand{\unb}{$^{\diamondsuit}$}

\newcommand{\langp}[2]{#1$\rightarrow$#2}

\newcommand{\repo}{\url{https://github.com/deep-spin/it-iwslt-2025}\xspace}

\usepackage[commandnameprefix=ifneeded, todonotes={textsize=small},final]{changes}
\definechangesauthor[color=magenta]{ga}
\newcommand{\ga}[1]{\added[id=ga]{#1}}

\title{Instituto de Telecomunicações at IWSLT 2025: Aligning Small-Scale Speech and Language Models for Speech-to-Text Learning}

\author{
 \textbf{Giuseppe Attanasio\IT},
 \textbf{Sonal Sannigrahi\IT \ist},
 \textbf{Ben Peters\IT},
 \textbf{André F.T. Martins\IT \ist \unb}
\\
 \IT Instituto de Telecomunicações, Lisbon, Portugal \\
  \ist Instituto Superior Técnico, Universidade de Lisboa, Portugal\\
  \unb Unbabel, Lisbon, Portugal \\
 \small{
     \href{mailto:giuseppe.attanasio@lx.it.pt}{\texttt{giuseppe.attanasio@lx.it.pt}}
 }
}

\begin{document}
\maketitle
\begin{abstract}
This paper presents Instituto de Telecomunicações's submission to the IWSLT 2025 Shared Task on Instruction Following Speech Processing.
We submit results for \ga{the Short Track, i.e., speech recognition, translation, and spoken question answering.} Our model is a unified speech-to-text model that integrates a \ga{pretrained continuous} speech encoder and text decoder through a first phase of modality alignment and a \ga{second} phase of instruction fine-tuning. Crucially, we focus on using \ga{small-scale language model backbones (< 2B) and restrict to high-quality, CC-BY data} along with synthetic data generation to supplement existing resources.\footnote{Code and data at \repo.}

\end{abstract}

\section{Introduction}

This paper presents our submission to the IWSLT 2025 Instruction Following track for the tasks of Automatic Speech Recognition (ASR), Speech Translation (ST), and Spoken Question Answering (SQA) for English, Chinese, and German.  
Our work builds upon a long line of previous research equipping LMs with additional multimodal capabilities, aligning an LM's semantic spaces with that of a pretrained speech encoder \cite[\it inter alia]{tang2023salmonn,huang2023speech,hu-etal-2024-wavllm,chu2024qwen2,grattafiori2024llama}.
Our contribution is particularly motivated by \textit{efficiency}, i.e., the goal of achieving strong performance using small-scale (< 2B) models. Recent work has explored audio quantization techniques \cite[\it inter alia]{zhang2024speechtokenizer, defossezhigh}, quantized input mel spectrograms \cite{shennaturalspeech}, or extreme compression of input data over the time dimension through convolutional kernels paired with strong small-scale LM backbones \cite{abouelenin2025phi}.

We take stock of such advancements and propose a model for even smaller scales. We use established methods for speech integration in LMs \citep{gaido-etal-2024-speech,grattafiori2024llama} using pretrained base models of up to 1.5B learnable parameters, finding empirically that with highly filtered and synthetic data, we can enable similar results at a fraction of the cost of larger LMs.
The main contributions of our system are as follows:
\begin{itemize}
    \item \textbf{Adapting pretrained, small-scale LMs:} We experiment with Qwen 2.5 1.5B and 0.5B \cite{qwen2.5} as our LM of choice and use w2v-BERT 2.0 \cite{barrault2023seamless} as our speech encoder.
    \item \textbf{Two-stage Training Curriculum:} We use a \textit{modality alignment} and \textit{instruction fine tuning} (IFT) phase for training our models, where the first equips the model with general speech capabilities and the second enables multi-task capabilities.
    \item \ga{\textbf{Training on open-licensed data:} To guarantee reproducibility and facilitate future research, we train on established CC-BY data collections and synthetic data filtered for quality. We release training and modeling artefacts under a permissive license.}
\end{itemize}

\section{Related Work}

\paragraph{Efficient LMs.} Recent works on efficient, small-scale language models (SLMs) have shown impressive knowledge compression capabilities by maintaining similar performance to larger, more computationally-intensive models.

Models such as Phi-4 Mini \cite{abouelenin2025phi} and Gemma 2 \cite{team2024gemma} have reported strong performance relative to size with a focus on computational efficiency. \citet{lu2024small} have shown how scaling laws operate differently for SLMs and have further demonstrated the efficiency of such models in subsequent reasoning tasks. 

\paragraph{Multimodal LM Extension.} Equipping a text-based model with multimodal capabilities is often done using an auxiliary modality encoder that is then used to jointly learn a semantic mapping between speech and text. Early works in joint text-speech modeling include AudioPaLM \cite{rubenstein2023audiopalm}, VioLA \cite{wang2023viola}, and VoxtLM \cite{maiti2024voxtlm}.
Other approaches combine a pretrained continuous speech encoder with an LM by concatenating speech embeddings to the text context \cite[\it inter alia]{tang2023salmonn,huang2023speech,hu-etal-2024-wavllm,chu2024qwen2,grattafiori2024llama}. 
Such works rely on strong multilingual capabilities of the speech encoder and those of large-scale LMs (i.e., 7B or more) to learn how to use speech-related parts of the context \citep{grattafiori2024llama}.
Our system echoes this compositional approach to speech and language modeling but leverages recent language models in the scale 0.5-1.5B.

\section{System Overview}

\subsection{Model Architecture}

Our model follows a standard speech encoder, text decoder architecture \cite[e.g.,][]{tang2023salmonn,grattafiori2024llama,chu2024qwen2}. 

\paragraph{Speech stack.}
We extract 80-dimensional Mel-filterbank audio representations with a stride of 2 using w2v-BERT 2.0's standard processor. Then we compress the audio over the time dimension using three 1D convolutional layers with a kernel width of 3 and a stride of 2. This input is then processed by the pretrained w2v-BERT 2.0 model. The output representations are processed by a modality and length adapter composed of two Conformer-like \citep{conformer} layers that further compress the audio representations on the time dimension and project them into the embedding space of the language model. 

\begin{figure}[!t]
    \centering
    \includegraphics[width=\linewidth]{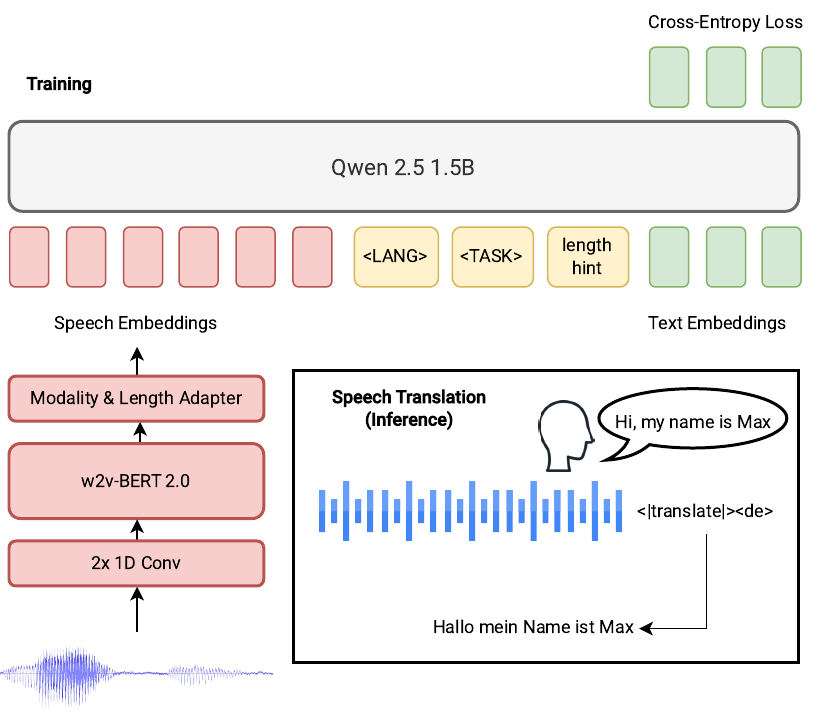}
    \caption{Illustration of our model. During training, the speech stack (red) generates speech representations which are prepended to task and language tags and an (optional) length hint (yellow), and text tokens (green). At inference time, we only provide the language and task tokens.}
    \label{fig:model_arch}
\end{figure}

\paragraph{Text stack.}
We prepend the audio representations computed from the audio stack to the text input embeddings extracted from the input embedding matrix of the language model. We use a bidirectional self-attention for the audio positions and a causal (autoregressive) one for the text part of the context. Following prior work \cite{chu2023qwen,Radford2022RobustSR}, we constrain text generation using a target language and task token.
Figure~\ref{fig:model_arch} illustrates the model architecture. 

\subsection{Training Curriculum}
 
We train our model in two stages: a modality alignment stage followed by an instruction fine-tuning (IFT) stage. 

\paragraph{Modality Alignment.} 
This first stage aligns the speech stack output representations to the language model's embedding space. 
We use the pretrained w2v-BERT 2.0 (v2)\footnote{\url{https://huggingface.co/facebook/w2v-bert-2.0}} as our speech encoder and randomly initialize the pre-encoder convolutional layers and the modality adapter.

We choose Qwen 2.5 1.5B,\footnote{\url{https://huggingface.co/Qwen/Qwen2.5-1.5B}} a multilingual LM, as our text decoder. Choosing a small (< 2B) model allows for the exploration of more efficient alternatives and is often overlooked in the literature.

In this phase, we train only the pre-encoder convolutional layers and the modality adapter with a learning rate of $3 \times 10^{-3}$ for a single epoch. We train the model only on ASR data. 
The model is trained using standard cross-entropy loss on the reference transcript tokens. With a 95\% chance, we prepend to the language and task tags a \textit{length hint}, as suggested by \citet{deitke2024molmo}, to let the model learn a length distribution.  
This stage leads to a model that can perform ASR but does not yet have other capabilities.

\paragraph{Instruction Fine-Tuning.} Following the modality alignment phase, we perform IFT using speech-to-text tasks included in the IWSLT campaign (AST, SQA) in English, German, and Chinese. During this stage, we train every component jointly end-to-end.\footnote{MA and IFT runs required a total of three days using four H100 GPUs in an in-house infrastructure.}

\paragraph{Generation Parameters.} For all tasks, we let the model generate up to 1024 tokens with beam search decoding (beam size of 3), a repetition penalty of 1.6, and nucleus sampling with temperature of 1.2. 

\subsection{Data}

Where possible, we use CC-BY licensed data across all tasks.
When sufficient data is not available, we generate synthetic corpora using the procedures described below. Table \ref{tab:data} provides an overview of the data sources used for each task and training phase.

\begin{table}[!t]
    \small
    \centering
    \begin{tabular}{lllr}
        \toprule
        \textbf{Task} & \textbf{Data} & \textbf{License} & \textbf{Hours} \\
        \midrule
        \multicolumn{4}{c}{Modality Alignment (MA)} \\
        \midrule
        \multirow{5}{*}{ASR} & LibriSpeech (LS) & CC-BY 4.0 & 1K \\ 
        & Multilingual LS &CC-BY 4.0 & 2K \\
        & FLEURS & CC-BY 4.0 & 24 \\
        & CommonVoice 16.1 & CC-BY 4.0 & 4K\\
        \midrule \midrule
        \multicolumn{4}{c}{Instruction Fine-Tuning (IFT)} \\
        \midrule
        \multirow{4}{*}{ASR} & \textit{All MA data} &  &  \\ 
        & VoxPopuli & CC-BY 4.0 & 1.8K \\ 
        & Peoples Speech & CC-BY 4.0 & 12K \\
        & CV 16.1 PL & - & 30K \\

        \multirow{2}{*}{ST} & CoVoST-2 & CC-BY NC 4.0 & 3K\\
        & CoVoST-2 PL & CC-BY NC 4.0 & 3K\\
        \multirow{2}{*}{SQA} & SpokenSQuAD & - & - \\ 
        & Generated Data & - & -\\        
         \bottomrule
    \end{tabular}
    \caption{Data statistics with licence, hours of speech data across all languages, and task splits.}
    \label{tab:data}
\end{table}

\paragraph{Speech Recognition.} We use CommonVoice 16.1 \citep{ardila-etal-2020-common}, FLEURS \citep{conneau2023fleurs}, MLS \citep{mls}, and LibriSpeech \citep{librispeech} for the modality alignment (MA) data mixture. For IFT, we reuse all MA data plus VoxPopuli  \citep{voxpopuli} and The People's Speech (clean) \citep{peoples}.

\paragraph{Speech Translation.}
We use CoVoST2 \citep{wang2020covost} as gold-standard AST data across English, Chinese, and German. We supplement this gold standard parallel data by \textbf{pseudolabeling} ASR transcriptions.
This technique has proven effective for previous systems \citep{barrault2023seamless,ambilduke2025tower} and is simple to implement.
Concretely, we translate all transcriptions from the English portion of CommonVoice 16.1 using four strong MT models: the 13B and Mistral-7B versions of TowerInstruct \citep{alves2024tower}, EuroLLM-9B-Instruct \citep{eurollm}, and NLLB-3.3B \citep{nllb-22}.
For each transcription, we use a COMETKiwi \citep{rei-etal-2022-cometkiwi} oracle to select the best translation among the four systems.
We then filter out examples for which the best translation records a score under $0.85$.
This process allows for fewer examples to be filtered than in conventional single-model pseudolabeling, as is shown in Table~\ref{tab:pseudolabels}, and also increases diversity because the translations come from a mixture of several models.

\begin{table}[!t]
    \small
    \centering
    \begin{tabular}{lrr}
        \toprule
        & \multicolumn{2}{c}{\% Kept} \\
        Model & \langp{en}{de} & \langp{en}{zh} \\
        \midrule
         NLLB-3B & 58.1 & 34.3\\
         TowerInstruct-Mistral-7B & 60.0 & 51.5\\
         TowerInstruct-13B & 59.4 & 49.9\\
         EuroLLM-9B-Instruct & 62.5 & 52.3 \\
         Oracle & 71.0 & 64.3\\
         \bottomrule
    \end{tabular}
    \caption{Percentage of English transcriptions in CommonVoice 16.1 for which various models produce a translation with a COMETKiwi score of at least $0.85$. The oracle keeps a much larger share of hypotheses than any individual model.}
    \label{tab:pseudolabels}
\end{table}

\paragraph{Spoken Question Answering.} We use the Spoken SQuAD \cite{lee2018spoken} dataset for English SQA. This dataset consists of several texts which are synthesized into speech and has a total of 37K questions and answers. Due to the limited availability of multilingual SQA datasets, we follow the same pseudolabeling process as for ST to create synthetic German and Chinese questions and answers for each example.
The question and answer were translated separately using the same mixture of models as for ST.
Question-answer pairs were kept if the best translated question had a COMETKiwi score of at least $0.80$.
The same system was used to translate the answer regardless of how it compared to the translated answers from other systems.\footnote{As the answers were generally very short, we found that COMETKiwi performed unreliably for them.}
As the SQA task also includes questions where the answer cannot be inferred from the context, we additionally generate synthetic \textit{unanswerable} questions for each context in English, German, and Chinese using Qwen2.5-70B \cite{qwen2.5}.
 
Following insights from \cite{sannigrahi2024synthetic}, we provide the LM with context along with example questions to guide the style and quality of the generated answers. We find that without example questions based on the original dataset, the LM often produces i) questions not adhering to the topic of the context and ii) verbose questions. We also experiment with prompts that do not explicitly request the model to mention aspects/topics of the context provided and find this to be suboptimal. As the number of positive instances in the Spoken QA dataset is small, in order to maintain a balanced dataset, we limit unanswerable questions to two per context. We further experimented with using the audio directly as opposed to the text transcript for context, but found this approach to be more prone to errors, as the Spoken SQuAD dataset is \textit{not} a native spoken dataset but rather a synthesized QA dataset, often leading to minor pronunciation errors. The final prompt used to obtain additional questions is shown in Figure \ref{fig:prompt}.

\begin{figure}[t]
\begin{minipage}[t]{\linewidth}
\begin{tcolorbox}[
colback=boxgray,
  colframe=gray,
  sharp corners,
  boxrule=0.8pt,
  top=6pt, bottom=6pt, left=5pt, right=5pt,
label={box:translation-prompt}
]

\small
    
Given a text passage and some questions about it, write 2 questions in [LANG\_ID] as close to the style of the original questions as possible but that are not answerable. The questions must be of similar difficulty as the example questions, i.e., they have to mention aspects and topics of the passage, but the answer cannot be inferred from the text. Be creative. Provide one question per line. \\

Text passage: [CONTEXT] \\

Example questions: [QUESTION] \\
Unanswerable questions: 

\end{tcolorbox}
\end{minipage}%

\caption{Prompt used to generate unanswerable questions from Qwen2.5-70B where \textbf{context} is the transcript used to synthesize speech in Spoken SQuAD, \textbf{question} is an answerable question from Spoken SQuAD, and \textbf{lang id} is the language in which we want to generate questions.}
\label{translation-prompt}
\label{fig:prompt}

\end{figure}

\paragraph{Preprocessing.}
We restrict our model to process audio of up to 120 seconds, discarding all training input longer than that. We preprocess all the instances appending to the speech embeddings (and prepending to the text embeddings) the task and language tags following the input template in Figure~\ref{fig:model_arch}. 

The task task tag can be either \texttt{<|transcribe|>}, \texttt{<|translate|>}, or \texttt{<|reply|>} for ASR, AST, and SQA, respectively, and the language tag is \texttt{<|en|>}, \texttt{<|de|>}, or \texttt{<|zh|>}.

\section{Results}

Our results for all three short-track tasks are in Table \ref{tab:results_iwslt}. For further details about the evaluation campaign as well as the metrics, we refer readers to \citet{abdulmumin-etal-2025-findings}. 

Our model obtains \textbf{reasonably good ASR scores for English}.
This result is particularly relevant, considering that the test data originate from the technical domain, exhibit high speaker variability, and consist primarily of spontaneous speech.

However, while the model successfully performs ASR, \textbf{SQA and ST prove to be more complex}. 
Through manual inspection, we observed poor quality outputs for ST. At times, the model repeats the same word or ignores the task tag and transcribes the audio segment rather than translating it.

This finding aligns with prior work that has found ASR data dominates the multitask capabilities of models \citep{tang2023salmonn}. Moreover, it emphasizes the importance of a more carefully designed training curriculum, where SQA, ST, and ASR data are more evenly distributed. 
Lastly, due to the audio length cutoff---set to 120 seconds due to technical limitations---we were unable to use all of the available SQA data. At test time, when prompted to perform the SQA task, the model sometimes generates the question itself, rather than the answer. We believe that by utilizing a combination of more data, enhanced base models with stronger multilingual capabilities, and extended context support, we will be able to improve upon these results substantially. 

\begin{table}
    \centering
    \small
\begin{tabular}{rr|rr|rr}
\toprule
  \multicolumn{2}{c|}{\textbf{en}} & \multicolumn{2}{c|}{\textbf{en-de}} & \multicolumn{2}{c}{\textbf{en-zh}} \\
  \midrule
  ASR & SQA & ST & SQA & ST & SQA \\ 
 \midrule

  0.15 & 0.14 & 0.34 & 0.22 & 0.34 & 0.21\\
\bottomrule
\end{tabular}
\caption{Official normalized ASR (WER ($\downarrow$)), ST (COMET ($\uparrow$)), and SQA (BertScore ($\uparrow$)) scores.}
\label{tab:results_iwslt}
\end{table}

\section{Conclusions}

We have presented our submission for the IWSLT 2025 Instruction Following Short Track. We explored the usage of a small-scale LM in modality adaptation through a continuous speech encoder. In particular, we equip an existing text model, Qwen 2.5 1.5B, with the speech modality for a joint multilingual and multitask model. 

We used standard modality alignment approaches, including building on pretrained speech encoders and autoregressive text decoder models, and a two-stage curriculum learning.

In future work, we plan to support longer contexts, better filtered data, and further push small-scale LMs to be fully multimodal. We will incorporate more high-quality multilingual data to enhance the model's language identification capabilities.
Additionally, we will extend the evaluation beyond standard performance-oriented benchmarks, e.g., by accounting for safety \citep{yang-etal-2024-towards-probing} and fairness \citep{koudounas2024towards,attanasio-etal-2024-twists}. 

\section*{Acknowledgements}
We thank Duarte Alves and Patrick Fernandes for their feedback and insightful discussions in the earlier versions of the paper.
This work was supported by the Portuguese Recovery and Resilience Plan through project C645008882-00000055 (Center for Responsible AI), by EU's Horizon Europe Research and Innovation Actions (UTTER, contract 101070631), by the project DECOLLAGE (ERC-2022-CoG 101088763), and by FCT/MECI through national funds and when applicable co-funded EU funds under UID/50008: Instituto de Telecomunicações. 

\section*{Limitations}

Currently, our model supports audio up to 2 minutes in length. Ideally, we would also like to support longer audio contexts while maintaining computationally inexpensive training. With the current length filters, we do not see much of the SQA data, which hinders the model's multi-task capabilities. Additionally, we have worked with a small LM (1.5B) in our model, which did not have the best language modeling capabilities. We plan to run additional experiments within the 3B scale. Lastly, there is limited research on the filtering of synthetically generated data for the QA domain. For future work, we plan to further refine the pipeline to generate synthetic QA data from spoken contexts.

\bibliography{anthology,custom}

\end{document}